\documentclass[11pt]{article} 
\usepackage{authblk}
\usepackage{amsthm}

\usepackage{booktabs}
\usepackage{longtable}
 
\usepackage{listings}
\usepackage{float}
\usepackage{xcolor}  
\usepackage{graphicx}  
\usepackage{amsmath}  
\usepackage{amssymb}  
\usepackage[linesnumbered,ruled,vlined]{algorithm2e}
\usepackage[
    style=numeric,
    sorting=none,        
    doi=false,
    isbn=false,
    url=false,
    eprint=false,
    language=auto
]{biblatex}
\DeclareBibliographyDriver{article}{%
  \printnames{author}%
  \setunit{\labelnamepunct}\newblock
  \printfield{title}%
  \setunit{\labelnamepunct}\newblock
  \printfield{journaltitle}%
  \setunit*{\addspace}%
  \printfield{volume}%
  \setunit*{\addspace}%
  \printfield{year}%
  \newunit
  \printfield{pages}%
  \finentry
}
\DeclareFieldFormat{pages}{pp.\ #1}
\DeclareBibliographyDriver{book}{%
  \printnames{author}%
  \setunit{\labelnamepunct}\newblock
  \printfield{title}%
  \newunit\newblock
  \printlist{publisher}%
  \setunit*{\addcomma\space}%
  \printlist{location}%
  \setunit*{\addcomma\space}%
  \printfield{year}%
  \finentry
}
\usepackage{float}
\usepackage{algorithmic}  
\usepackage{placeins} 
\usepackage{hyperref}  
\usepackage{titlesec}  
\usepackage{appendix}  
\usepackage{subcaption}
\usepackage{setspace}  
\usepackage[margin=1.2in]{geometry}
\makeatletter
\renewenvironment{abstract}
  {\leftskip\z@ \rightskip\z@ \small}
  {\par\vspace{1em}}
\makeatother

\title{\textbf{From Theory to Practice with RAVEN-UCB: Addressing Non-Stationarity in Multi-Armed Bandits through Variance Adaptation}}
\author[1]{Junyi Fang\thanks{Corresponding author. Email: \href{mailto:junyifang@csu.edu.cn}{junyifang@csu.edu.cn}}}
\author[1]{Yuxun Chen}
\author[2]{Yuxin Chen}
\author[3]{Chen Zhang}

\affil[1]{School of Business, Central South University, Changsha 410083, China}
\affil[2]{School of Finance and Statistics, Hunan University, Changsha 410006, China}
\affil[3]{Wisdom Lake Academy of Pharmacy, Xi’an Jiaotong-Liverpool University, Suzhou 215123, Jiangsu, China}
\date{}

\begin{document}
\maketitle
\section*{Abstract}
\begin{abstract}
The Multi-Armed Bandit (MAB) problem faces significant challenges in non-stationary environments where reward distributions dynamically evolve. We propose RAVEN-UCB, a novel bandit algorithm that bridges theoretical rigor with practical efficiency through variance-aware adaptation. Theoretically, RAVEN-UCB achieves tighter regret bounds than UCB1 and UCB-V, with gap-dependent regret $\mathcal{O}\left(\frac{K\sigma_{\max}^{2}\log T}{\Delta}\right)$ and gap-independent regret $\mathcal{O}\left(\sqrt{KT\log T}\right)$. Practically, it integrates three key innovations: (1) Variance-driven exploration via $\sqrt{\hat{\sigma}^2_k/(N_k+1)}$ in confidence bounds, (2) Adaptive control through $\alpha_t = \alpha_0 / \log(t + \epsilon)$, and (3) $\mathcal{O}(1)$ recursive updates for computational efficiency. We validate RAVEN-UCB through a series of experiments across diverse non-stationary patterns—distributional parameter changes, periodic shifts, and temporary fluctuations—using both synthetic environments and a large-scale logistics case study. Results demonstrate consistent superiority over state-of-the-art baselines, confirming the theoretical advantages while highlighting practical robustness in dynamic decision-making scenarios.
\end{abstract}

\section*{Keywords}
Reinforcement learning, variance-based methods, Non-Stationary MAB, exploration-exploitation balance, UCB

\section{Introduction}

The Multi-Armed Bandit (MAB) problem is a key concept in reinforcement learning and decision theory.  An agent sequentially selects among multiple actions, or ``arms,'' to maximize cumulative rewards. Each arm gives random rewards from an unknown distribution.  The goal is to get as many rewards as you can by balancing \emph{exploration} (learning about arms' reward distributions) and \emph{exploitation} (choosing the arm with the highest estimated reward)\cite{liu_modification_2016,auer_ucb_2010}. This framework, first formalized by Robbins~\cite{robbins1952some}, has led to big changes in areas like online advertising~\cite{li2010contextual}, recommendation systems~\cite{li2010contextual, li2011unbiased}, and logistics optimization~\cite{chen2013combinatorial}. Algorithms such as Upper Confidence Bound (UCB), $\epsilon$-greedy, and Thompson Sampling have been studied a lot. They have shown good performance in stationary environments, where reward distributions stay constant over time~\cite{lattimore2020bandit}. However, many real-world situations are different from this assumption. They show \textit{non-stationarity}, where reward distributions $D_k(t)$ evolve dynamically due to things like changing user preferences, market fluctuations, or operational disruptions~\cite{besbes2014stochastic}. This is similar to \textit{concept drift} and shows the need for adaptive MAB algorithms to manage the exploration-exploitation trade-off in dynamic environments. The development of these algorithms is essential for two main reasons. First, they improve performance in current applications. Second, they allow new use cases in environments that are unpredictable. This makes decision-making systems more robust and reliable. 

There has been progress in dealing with non-stationarity in MAB problems. But, current approaches have big limits that make them less effective in dynamic settings.  We can divide these methods into two  main groups. One group uses sliding windows or discounting to deal with mean drift. The other group includes variance information in a static way.  Sliding-window methods like Sliding-Window UCB (SW-UCB)~\cite{garivier2011upper} use a fixed-size window of recent observations to estimate reward distributions. Discounting techniques like Discounted UCB (D-UCB)~\cite{kocsis2006discounted} use exponential weights to prioritize recent data. These approaches can follow changes in reward means. But they often neglect variance dynamics ($\sigma^2_k(t)$). Variance dynamics are important in environments with changing uncertainty. Recently, algorithms like f-Discounted-Sliding-Window Thompson Sampling (f-dsw TS)~\cite{cavenaghi_non_2021} have been proposed to address these shortcomings. These algorithms combine sliding-window and discounting strategies with dynamic variance adaptation. This helps improve adaptability in non-stationary settings. 

Second, variance-aware algorithms like UCB-V~\cite{audibert_explorationexploitation_2009} use static variance estimates. They do not capture temporal fluctuations and limiting dynamic exploration. In contrast, RAVEN-UCB employs more flexible parameters with a time-decaying exploration factor. It selects better arms less often, so it is more effective in dynamic environments. 
Furthermore, sliding-window and variance-aware methods have computational inefficiencies. Maintaining windows or recalculating statistics from scratch incurs $O(n)$ time complexity per update~\cite{bubeck_regret_2012}, which is impractical for large-scale, real-time applications like online advertising or logistics. To address these gaps, we propose RAVEN-UCB, a new variance-adaptive bandit algorithm  for non-stationary environments. RAVEN-UCB introduces three key contributions:

\begin{enumerate}
    \item \textbf{Variance as an Exploration Signal}: Unlike traditional methods, RAVEN-UCB dynamically adjusts exploration based on real-time sample variance, incorporating a term proportional to $\sqrt{\frac{\hat{\sigma}^2_k}{N_k + 1}}$ in its upper confidence bound, where $\hat{\sigma}^2_k$ is the estimated variance of arm $k$ and $N_k$ is the number of times it has been selected (Sec. \ref{Algorithm Implementation}). This enables increased exploration during periods of high variance, reflecting greater uncertainty or potential distribution shifts.
   \item \textbf{Flexible and Robust Parameterization}: RAVEN-UCB uses a highly adaptable parameterization with a time-changing exploration coefficient $\alpha_t = \alpha_0 / \log(t + \epsilon)$ and adjustable parameters $\beta_0$ and $\epsilon$. This design ensures responsiveness to environmental changes while maintaining low sensitivity to parameter choices(Sec. \ref{Parameter Selection}).It ensures stable performance in different non-stationary situations.
    \item \textbf{Efficient Recursive Updates}: To enhance scalability, RAVEN-UCB uses recursive formulas for updating sample mean and variance, achieving $O(1)$ time complexity per step (Eq. \ref{eq:recursive_mean}--\ref{eq:recursive_variance}). This overcomes the computational bottlenecks of naive method, making it suitable for large-scale applications.
\end{enumerate}

We  organize the remainder of this paper as follows: Section \ref{Background} provides a comprehensive background on the MAB problem, formalizing non-stationary environments through classes such as Distributional Parameter Changes (DPC), Periodic Changes (PC), and Temporary Fluctuations (TF), and reviews existing adaptive bandit algorithms. Section \ref{Methodology} presents the RAVEN-UCB algorithm. We detail its design principles, variance-adaptive exploration, logarithmic decay mechanism, and recursive update formulas, along with theoretical regret bound analysis. Section \ref{Experiments} evaluates RAVEN-UCB’s empirical performance through three experiments: a regret comparison against UCB1, showing an average regret reduction of 84\% compared to UCB1; a sensitivity analysis of hyperparameters across different scenarios; and a simulated logistics optimization case study with 100 warehouses, where RAVEN-UCB achieves 68\% lower regret than standard UCB. Section \ref{Conclusion} concludes by summarizing the main contributions and outlining future research directions, including extensions to contextual bandits and integration with large language models for enhanced decision-making. Detailed derivations and proofs are provided in the Appendix \ref{app:Appendix}.

\section{Background}
\label{Background}
  \subsection{Multi-Armed Bandit Problem}
  
Since it was first introduced in the 1950s by Robbins~\cite{robbins1952some}, the Multi-Armed Bandit Problem has been established as a fundamental framework for sequential decision-making under uncertainty. In a typical MAB setting, an agent selects one of \( K \) arms at each time step \( t \), receiving a reward \( r_k(t) \sim D_k(t) \). And \( D_k(t) \) is the reward distribution for arm \( k \)\cite{cavenaghi_non_2021}. Different policies have been developed to determine which arm to select at each time step in the multi-armed bandit problem. The most studied in the scientific researches are:
\begin{itemize}
    \item \textbf{Upper Confidence Bound (UCB):} This Algorithm picks the arm with the highest sum of the estimated mean reward and an uncertainty term, scaled by a parameter \(\alpha\) to control exploration versus exploitation. The chosen arm \(a\) is determined as \( a \leftarrow \underset{k \in \mathcal{K}}{\operatorname{argmax}} \left\{ \hat{\mu}_k(t) + \alpha \cdot f(\hat{\sigma}_k(t)) \right\} \), where \(\hat{\mu}_k(t)\) is the estimated mean reward for arm \(k\) at time \(t\), \(\hat{\sigma}_k(t)\) is the estimated standard deviation, and \(f\) scales the standard deviation.\cite{auer2002finite}
    \item \textbf{\(\varepsilon\)-greedy:} This strategy explores by randomly selecting an arm with probability \(\varepsilon\), while exploiting the arm with the highest estimated mean reward with probability \(1 - \varepsilon\).\cite{tokic2010adaptive,fayyazi_real-time_2023}
    \item \textbf{Thompson Sampling:} A Bayesian method that maintains a posterior distribution for each arm’s mean reward, sampling from these distributions at each step and selecting the arm with the highest sampled value.\cite{Byrd2023}
\end{itemize}

The methods above have been extensively analyzed, and lots of theoretical results guarantee their convergence (i.e., regret bound) to the optimal solution in stationary environments.~\cite{audibert_explorationexploitation_2009,mukherjee2018efficient}
A stationary setting is defined as an environment in which the reward distribution \( D_k \) for each arm \( k \) is assumed to be stationary does not change through all the time-steps in \( T \).\cite{cavenaghi_non_2021}

MAB algorithms are used in many areas. Important application areas include online advertising, recommendation systems, and logistics and supply chain optimization. In online advertising, MAB algorithms help to optimize ad selection in real time to maximize click-through rates and conversion rates. For instance, Jahanbakhsh et al.\cite{DBLP:journals/corr/abs-2011-10919} modeled ad selection as a MAB problem, dynamically identifying high-performing ads using online learning techniques. Similarly, Nguyen-Thanh et al.\cite{DBLP:journals/corr/abs-1909-04190} proposed a UCB-based recommendation strategy that addresses large action spaces and non-stationary user preferences, significantly improving user engagement. In recommendation systems, MAB algorithms address challenges such as the cold-start problem and dynamically evolving user preferences\cite{li2024improvement}. Ding et al.\cite{xia2024spotify} applied $\epsilon$-greedy, Thompson Sampling, and UCB algorithms to personalize product recommendations on an e-commerce platform, achieving notable improvements in user interaction metrics.
Meanwhile, in logistics and supply chain management, MAB algorithms are applied to dynamic pricing, inventory control, and resource allocation. Gao and Zhang\cite{gao2022efficient} developed a UCB-based learning framework to better customer selection in multi-product inventory systems, achieving efficient stock management. In emergency logistics, geometric greedy algorithms were proposed to optimize hub locations and resources distribution under uncertainty, improving how quickly and strongly supply chains respond~\cite{liang2024multi}.

   \subsection{Non-Stationarity in MAB}
The MAB problem is a basic idea for making decisions when things are uncertain, and lots of research looks at how it can be used in different situations.  Real-world uses often differ from basic ideas. For instance, Wang et al.\cite{gulcu2021multi} extend the MAB framework to scenarios where the optimal arm depends on a hidden Markov model (HMM), introducing reward correlations governed by an unknown Markov process. Wang, Wang, and Huang address the challenge of combined and anonymous feedback, where rewards are delayed and mixed across actions, like online advertising \cite{wang_adaptive_2021}. They introduced adaptive algorithms (ARS-UCB and ARS-EXP3) to get the best regret bounds without prior knowledge of delay structures, making MAB stronger in both predictable and challenging settings. These improvements show how MAB can adjust to tricky environments, dealing with things like reward links and feedback waits. 
In real-world situations, such as logistics, reward distributions often change over time due to factors like traffic conditions, inventory levels, or user preferences, leading to non-stationary MAB problems\cite{cavenaghi_non_2021}. This non-stationarity is similar to \emph{concept drift} in machine learning where the conditional distribution \( P(y|X) \) changes\cite{kuncheva2008classifier}. This challenges traditional algorithms that rely on fixed distributions. Non-stationary MAB environments are divided based on how reward distributions \( D_k(t) \), defined by mean \( \mu_k(t) \) and variance \( \sigma_k^2(t) \), evolve over time. These are further broken down into three main types:

\begin{itemize}
    \item \textbf{Distributional Parameter Changes(DPC):} The mean or variance changes over time:
    \[
    D_k(t) = D(\mu_k(t), \sigma_k^2(t)),
    \]
    where \( \mu_k(t) = \mu_k(0) + \delta_k(t) \), \( \sigma_k^2(t) = \sigma_k^2(0) + g_k(t) \), and \( g_k(t) \geq 0 \). 
    \item \textbf{Periodic Changes(PC):} The distribution repeats with period \( P \):
    \[
    D_k(t) = D_k(t\mod P).
    \]
    \item \textbf{Temporary Fluctuations(TF):} The distribution deviates briefly and reverts:
    \[
    D_k(t) = 
    \begin{cases} 
    D_k^{\text{normal}}, & t < t_b \text{ or } t \geq t_b + \Delta t, \\
    D_k^{\text{blip}}, & t_b \leq t < t_b + \Delta t,
    \end{cases}
    \]

\end{itemize}
Meanwhile, the following table \ref{tab:ns_mab_reclass_en} summarizes the specific real-world manifestations, mathematical abstractions, and examples of several Non-stationary MAB environments.
\begin{table}[htbp]
\centering
\caption{Some Non-Stationary MAB Scenarios}
\label{tab:ns_mab_reclass_en}
\scriptsize
\setlength{\tabcolsep}{4pt}
\begin{tabular}{@{}p{3cm}|c|p{6cm}|p{6cm}@{}}
\toprule
\textbf{Scenario} & \textbf{Cat.} & \textbf{Math. Definition} & \textbf{Real-World Example} \\

\midrule
Incremental Drift 
 & DPC 
 & \(\mu_k(t)=\mu_k(0)+\delta_k\,t,\;\delta_k\ll1\) 
 & User preference evolves slowly \\ 
\midrule
Variance Drift 
 & DPC 
 & \(\sigma_k^2(t)=\sigma_k^2(0)+g_k(t),\;g_k(t)\ge0\) 
 & Traffic variability affects delays \\ 
\midrule
Gradual Drift 
 & DPC
 & 
 \(\displaystyle
 D_k(t)=
 \begin{cases}
   D_k^{\text{old}}, & t<t_0,\\
   D_k^{\text{old}} \text{ w.p. } 1-\rho(t); \\
   D_k^{\text{new}} \text{ w.p. } \rho(t), & t_0 \le t < t_1,\\
   D_k^{\text{new}}, & t \ge t_1
 \end{cases}
 \)
 & Gradual user migration from old to new \newline product version. \newline\\
\midrule

Localized Jump Drift 
 & DPC 
 & 
 \(\displaystyle
 D_k(t)=
 \begin{cases}
   D_k(t-1), & k \notin \mathcal{S}(t), \\
   \mathcal{U}(\mu_{\min}, \mu_{\max}), \\
   \quad \mathcal{U}(\sigma^2_{\min}, \sigma^2_{\max}), & k \in \mathcal{S}(t)
 \end{cases}
 \)
 & Edge device resets after network reconnection or resource change. \newline \\

\midrule
Periodic Drift 
 & PC  
 & \(D_k(t+P)=D_k(t)\) 
 & Seasonal demand cycles \\ 
\midrule
Blips / Outliers 
 & TF  
 & 
 \(\displaystyle
 D_k(t)=
 \begin{cases}
   D_k^{\mathrm{normal}}, & t\notin[t_b,t_b+\Delta t),\\
   D_k^{\mathrm{blip}},   & t\in[t_b,t_b+\Delta t)
 \end{cases}
 \) 
 & Short-term strike / outage \\ 
\midrule
Add/Remove Arm
 & TF  
 & 
 \(\displaystyle
 \mathcal K(t)=
 \begin{cases}
   \mathcal K_0, & t<t_a,\\
   \mathcal K_0\cup\{k'\}, & t_a\le t<t_r,\\
   \mathcal K_0\cup\{k'\}\setminus\{k''\}, & t\ge t_r
 \end{cases}
 \) 
 & New warehouse opens / closes \\
\bottomrule
\end{tabular}
\end{table}

For example, in logistics, parameter changes might show efficiency variations due to traffic. Periodic changes may link to seasonal demand. Temporary fluctuations can mean short-term disruptions like  strikes \cite{ivanov2019impact}. These types show why adaptive strategies  are important for handling non-stationary MAB problems. Recent progress has been exploring large language models (LLMs), such as GPT-3, said Brown et al.\cite{brown2020language}, to offer new ideas on decision-making in dynamic environments. Discounted UCB (D-UCB) and Sliding-Window UCB (SW-UCB) reduce the impact of old rewards to handle drift\cite{garivier2011upper}. Similarly, there are discounted or windowed types of Thompson Sampling for sub-Gaussian rewards. Meanwhile, the f-Discounted-Sliding-Window Thompson Sampling (f-dsw TS) algorithm uses discount factors and sliding windows to enhance adaptability to concept drift \cite{cavenaghi_non_2021}. This approach builds on earlier adaptive strategies. It offers a strong solution for dynamic reward distributions in complex real-world situations. The classic epsilon-decreasing method exists, and SoftMax algorithm,   by Velonis and Vergos\cite{velonis_comparison_2022}, uses a probability distribution to adjust the likelihood of selecting each arm. It is based on its estimated reward, while sliding-window methods focus on recent data \cite{garivier2011upper}. Again, variance-aware methods like UCB-V guide exploration using variance estimates\cite{audibert_explorationexploitation_2009}.The limits of current methods and the importance of variance in understanding uncertainty push us to propose RAVEN-UCB.  This is a variance-adaptive algorithm.  It estimates mean and variance and adjusts exploration over time. This approach improves adaptability in changing environments, especially when distribution parameters change.

\section{Methodology}
\label{Methodology}
In this section, we show why and how the  \textbf{RAVEN-UCB} algorithm works. It builds on the classical Upper Confidence Bound (UCB) method and uses variance-based exploration to improve the approach. 


\subsection{Exploration Signal Based on Variance}

Variance helps measure uncertainty in random  variable distributions.~\cite{sheng2014some} When sampling is low, the sample mean and variance can change a lot.  These changes show where the agent can explore to get better reward estimates. So, the agent uses changes in variance as a signal to explore. 

For a given arm $k$, the sample mean $\hat{\mu}_k$ is calculated as:
\begin{equation}
\hat{\mu}_k = \frac{1}{n_k} \sum_{i=1}^{n_k} X_{k,i}
\label{eq:sample_mean}
\end{equation}

where \( X_{k,i} \) represents the reward from the \( i \)-th pull of arm \( k \), and \( n_k \) is the number of times arm \( k \) has been pulled.\cite{kun_upper_2024} The sample variance \( \hat{\sigma}^2_k \) is given by:

\begin{equation}
\hat{\sigma}^2_k = \frac{1}{n_k - 1} \sum_{i=1}^{n_k} (X_{k,i} - \hat{\mu}_k)^2
\label{eq:sample_variance_6666}
\end{equation}

where \( \hat{\sigma}^2_k \) estimates the variance of the reward distribution for arm \( k \).

When the sample size is small,  the sample mean and sample variance often fluctuate more significantly due to limited data. As the number of samples increases, the sample mean and variance become more stable, converging to the true population values:

\begin{equation}
\mu_k = \lim_{n_k \to \infty} \hat{\mu}_k \quad \text{and} \quad \sigma_k^2 = \lim_{n_k \to \infty} \hat{\sigma}^2_k
\label{eq:convergence}
\end{equation}

\begin{figure}[H]
\centering
\begin{subfigure}[t]{0.48\textwidth}
    \centering
    \includegraphics[width=\textwidth]{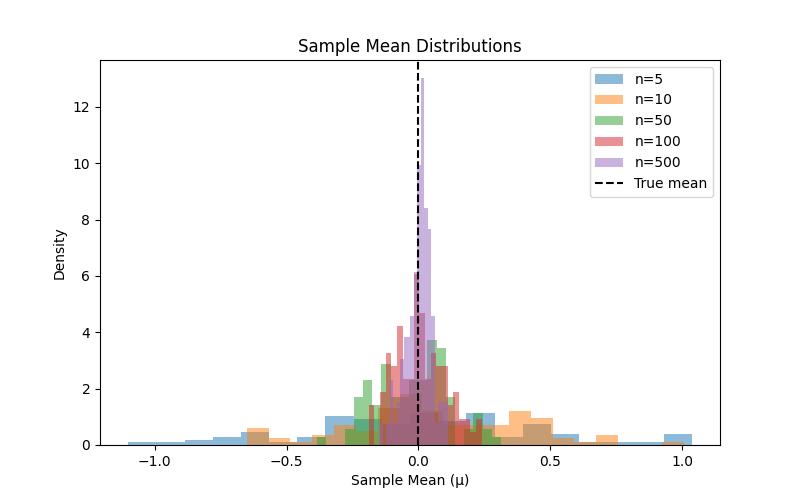}
    \caption{Distribution of sample means ($\hat{\mu}_k$) for a bandit arm with true mean $\mu_k = 0$, computed using Equation~\ref{eq:sample_mean} across 100 trials for various sample sizes $n_k$.}
    \label{fig:sample_mean_dist}
\end{subfigure}
\hfill
\begin{subfigure}[t]{0.48\textwidth}
    \centering
    \includegraphics[width=\textwidth]{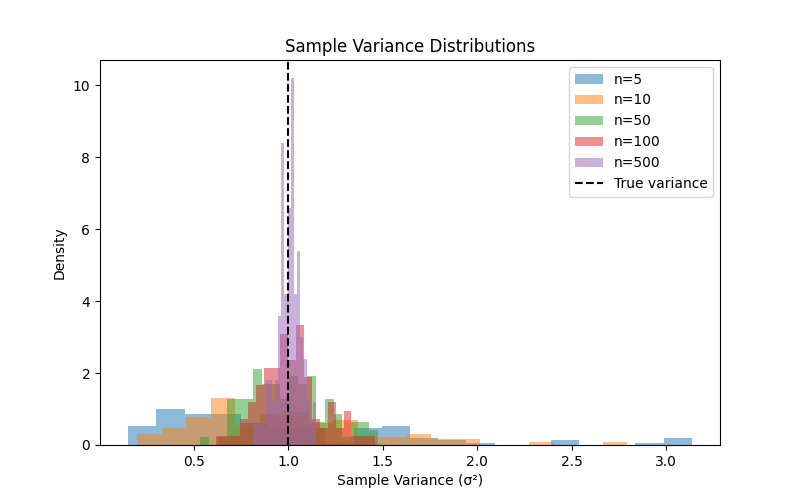}
    \caption{Distribution of sample variances ($\hat{\sigma}^2_k$) for a bandit arm with true variance $\sigma^2_k = 1$, computed using Equation~\ref{eq:sample_variance_6666} across 100 trials for various sample sizes $n_k$. }
    \label{fig:sample_variance_dist}
\end{subfigure}
\caption{Distributions of sample means and variances for a bandit arm}
\label{fig:mean_variance_dist}
\end{figure}

Thus, as more samples are collected, the variance estimates improve. This makes exploration focus on areas with more uncertainty. In our algorithm, variance changes are used as an exploration signal. This concept can be further clarified by examining a situation where a bandit arm has been sampled only a limited number of times. Under such conditions, the sample mean may not reliably approximate the true expected reward, and the corresponding sample variance is typically large, indicating significant uncertainty about the reward distribution of arm. As more samples are collected, the variance lowers and becomes stable. This gives a clearer picture of the arm's reward characteristics. This stabilization enhances the agent's confidence in deciding the best arm. To illustrate this, Figure~\ref{fig:sample_mean_dist} shows the distribution of sample means (Equation~\ref{eq:sample_mean}) for different sample sizes, demonstrating larger fluctuations for smaller $n_k$. Similarly, Figure~\ref{fig:sample_variance_dist} presents the distribution of sample variances (Equation~\ref{eq:sample_variance_6666}).
From a statistical point, we can see the relationship between the sample size and the variance. With a smaller sample size, the variance of the samples tends to be higher. This means the reward estimate is less reliable. The reason is not enough data to form a stable estimate. This phenomenon can be explained by the \textit{Law of Large Numbers}. As we get more observations, sample statistics, like mean and variance, get closer to the actual population statistics. \cite{baum1965convergence,judd1985law}

\subsection{Decaying Coefficient Design for Exploration Control }

We think that the classical Upper Confidence Bound (UCB) approach should add a variance-based exploration term with a decaying coefficient, i.e. \(\alpha_t = \alpha_0 / \log(t + \epsilon)\). This matches recent research highlighting its role in effective exploration strategies. Djallel Bouneffouf and Irina Rish.\cite{DBLP:journals/corr/abs-1904-10040} demonstrated that such adaptive exploration strategies with time-dependent decay are effective across healthcare, finance, and recommendation systems, particularly where exploration costs vary across actions. In the RAVEN-UCB algorithm, the exploration coefficient $\alpha_t$ is designed to decay logarithmically over time, specifically as $\alpha_t = \frac{\alpha_0}{\log(t + \epsilon)}$. This decay helps fix over-exploration problems  in standard UCB algorithms.  The exploration term $\sqrt{\frac{\ln t}{N(k)}}$ can lead to persistent sampling of suboptimal arms even after sufficient information has been gathered. By reducing $\alpha_t$ over time, the algorithm focuses from exploration to exploitation, enhancing convergence speed and stability. This is particularly beneficial in non-stationary environments, where reward distributions may change over time. The decaying term lets the algorithm adjust to changes better, balancing exploring new choices and using good ones.  Studies show that handling changes with techniques like decaying exploration or sliding windows improves bandit algorithms in dynamic settings. 
\newtheorem{proposition}{Proposition}

\subsection{Recursive Calculation of Sample Mean and Variance}
In multi-armed bandit experiments, as the number of samples increases, the sample mean (Equation~\ref{eq:sample_mean}) and sample variance (Equation~\ref{eq:sample_variance_6666}) change continuously. Traditionally, recalculating these statistics means storing and processing all past rewards for each arm at every step. This needs a lot of computational power and time.  This leads to a time complexity of $\mathcal{O}(n)$, where \(n\) is the number of samples. This approach becomes inefficient when computational efficiency is paramount.

The sample mean and variance naturally use information from all past data. This allows us to refine them step by step without starting over each time. Using this idea, we make recursive formulas for the sample mean and variance from a series of random variables. This lets us update them quickly at each step without checking old data again. Here are the recursive formulas:

\begin{proposition}[Recursive Formula for Sample Mean and Variance]
Given a sequence of random variables \( X_1, X_2, \dots, X_n \), the recursive formulas for the sample mean and variance are given by:
\begin{equation}
\overline{X}_{n+1} = \overline{X}_n + \frac{X_{n+1} - \overline{X}_n}{n+1}
\label{eq:recursive_mean}
\end{equation}
\begin{equation}
S_{n+1}^2 = \left( 1 - \frac{1}{n} \right) S_n^2 + (n+1) \left( \overline{X}_{n+1} - \overline{X}_n \right)^2
\label{eq:recursive_variance}
\end{equation}
\end{proposition}

The detailed derivation of these recursive formulas is provided in Appendix~\ref{app:recursive_derivation}.

\subsection{RAVEN-UCB Algorithm}
\label{Algorithm Implementation}
The RAVEN-UCB algorithm integrates the recursive calculations of  the sample mean and variance to dynamically adjust the exploration-exploitation trade-off. The algorithm is outlined in Algorithm~\ref{alg:v_ucb}, where the scores for each arm are computed based on the sample mean, exploration term, and variance term.
As established by Auer et al. in UCB1\cite{auer2002finite}, we add this coefficient($\alpha_0$) reduces exploration intensity over time while leveraging marginal changes in sample variance to guide arm selection, especially when reward distributions vary significantly.
\begin{algorithm}
\caption{RAVEN-UCB Algorithm}
\label{alg:v_ucb}
Initialize $N(k) = 0$, $M(k) = 0$, $S^2(k) = 0$ for $k = 1$ to $K$\;
Set parameters: $\alpha_0$, $\beta_0$, $\epsilon$, $T$\;
\For{each time step $t = 1$ to $T$}{
    \eIf{$t \leq K$}{
        $k_t \leftarrow t$\;
    }{
        Compute $\alpha_t = \alpha_0 / \log(t + \epsilon)$\;
        Compute scores for each arm $k$:\
        \[
        \text{score}(k) = M(k) + \alpha_t \cdot \sqrt{\frac{\ln(t + 1)}{N(k) + 1}} + \beta_0 \cdot \sqrt{\frac{S^2(k)}{N(k) + 1} + \epsilon}
        \]\
        $k_t \leftarrow \arg\max_k \text{score}(k)$\
    }
    Obtain reward $R_t \sim \text{Distribution}(\mu_{k_t}, \sigma_{k_t})$\;
    Update $N(k_t) \leftarrow N(k_t) + 1$\;
    $n \leftarrow N(k_t)$\;
    $M(k_t) \leftarrow M(k_t) + \frac{R_t - M(k_t)}{n}$\;
    \eIf{$n > 1$}{
        $S^2(k_t) \leftarrow S^2(k_t) + (R_t - M(k_{t-1})) \cdot (R_t - M(k_t))$\;
        $S^2(k_t) \leftarrow \frac{S^2(k_t)}{n - 1}$\;
    }{
        $S^2(k_t) \leftarrow 0$\;
    }
}
Output total reward and regret\;
\end{algorithm}

We have theoretically derived the regret bounds of the RAVEN-UCB algorithm (proof see Appendix~\ref{app:regret_bound}).  

\begin{itemize}
    \item \textbf{Gap-dependent regret:}
\begin{equation}
        R(T) = \mathcal{O}\left( \frac{K\sigma_{\max}^2 \log T}{\Delta} \right),
        \label{eq:gap_dependent_regret}
    \end{equation}
    
    \item \textbf{Gap-independent regret:}
    \begin{equation}
        R(T) = \mathcal{O}\left( \sqrt{KT \log T} \right).
        \label{eq:gap_independent_regret}
    \end{equation}
\end{itemize}
We summarize the theoretical regret bounds of RAVEN-UCB and several baseline algorithms in Table~\ref{tab:regret_comparison}.
\begin{table}[h]
\centering
\caption{Comparison of Regret Upper Bounds for Different Algorithms}
\label{tab:regret_comparison}
\begin{tabular}{|c|c|c|}
\hline
\textbf{Algorithm} & \textbf{Gap-Dependent} & \textbf{Gap-Independent} \\
\hline
\textbf{RAVEN-UCB (ours)} & $\mathcal{O}\left( \frac{K\sigma_{\max}^2 \log T}{\Delta} \right)$ & $\mathcal{O}\left( \sqrt{KT \log T} \right)$ \\
UCB1\cite{auer2002finite} & $\mathcal{O}\left( \frac{K\log T}{\Delta} \right)$ & $\mathcal{O}\left( \sqrt{KT\log T} \right)$ \\
UCBV\cite{audibert_explorationexploitation_2009} & $ \mathcal{O}\left( \sum_{k \neq k^*} \left( \frac{\sigma_k^2 \log T}{\Delta_k} + \log T \right) \right).$ & $\mathcal{O}\left( \sqrt{KT\log T} \right)$ \\
\hline
\end{tabular}
\end{table}

\section{Experiments}
\label{Experiments}
In this section, we evaluate the performance of the RAVEN-UCB algorithm through three experiments. First, we analyze its regret performance under sub-Gaussian rewards and compare it with classical bandit algorithms to see if our theoretical proofs hold. Next, we conduct an parameter selection study to assess the impact of key parameters and present guide for practical values recommendation. Finally, we compare RAVEN-UCB with advanced algorithms in a simulated logistics scenario.It demonstrates its effectiveness in minimizing regret under dynamic conditions.

\subsection{Regret Experiment}
To check the theoretical properties of the proposed RAVEN-UCB algorithm, we conducted extensive simulations comparing its performance with the classical UCB1 baseline. Our experiments address two key questions:

(1) How does variance improve regret performance compared to UCB1 ?  

(2) How does the regret reduction scale with \(T\) ? 

We set \(K=10\) arms with Bernoulli rewards, where the true means \(\theta_k\) are drawn uniformly from \([0.8,0.95]\). UCB1 serves as our baseline algorithm. We measure normalized regret reduction:
\begin{equation}
\frac{R_{\mathrm{UCB1}} - R_{\mathrm{V\text{-}UCB}}}{R_{\mathrm{UCB1}}} \times 100\%
\end{equation}

Hyperparameters \((\alpha_0,\beta_0,\epsilon)\) for RAVEN-UCB are tuned via Optuna~\cite{akiba2019optuna} over \(M=50\) trials per configuration, with search ranges
\(\alpha_0\in[0.01,10]\), \(\beta_0\in[0.01,10]\), and \(\epsilon\in[10^{-3},0.5]\).

\begin{figure}[h]
  \centering
  \includegraphics[width=0.8\textwidth]{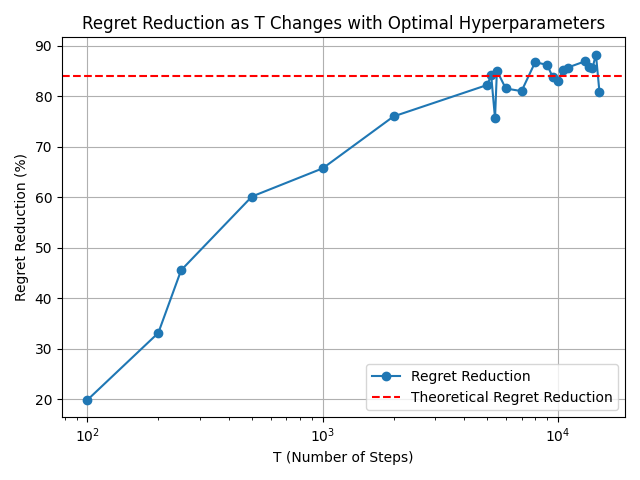}
  \caption{Regret reduction in horizon \(T\)}
  \label{fig:regret_scaling}
\end{figure}

Figure~\ref{fig:regret_scaling} shows how regret reduction changes as \(T\) increases.As \(T\) grows large (e.g.\ \(T\approx5000\)), the empirical variance estimates in RAVEN-UCB stabilize. This enables the algorithm to concentrate exploration on genuinely uncertain arms.  Empirically, we observe regret reduction is all above 80\%, matching the theoretical gap-dependent bound.

\FloatBarrier
Both algorithms pull each suboptimal arm \(i\) on the order of \(\log T\) times, but with different scaling factors dependent on reward variance:  
UCB1 selects suboptimal arms with a rate proportional to \(\frac{\log T}{\Delta_i^2}\), our RAVEN-UCB reduces this rate to \(\frac{\sigma_i^2 \log T}{\Delta_i^2}\) by incorporating variance exploration (See (\ref{eq:gap_dep}) in \ref{app:regret_bound}). The difference in their pull counts is therefore:

\begin{equation}
\left(\frac{\log T}{\Delta_i^2} - \frac{\sigma_i^2 \log T}{\Delta_i^2}\right)
\;=\;\frac{(1-\sigma_i^2)\log T}{\Delta_i^2},
\end{equation}

which grows linearly in \(\log T\). As \(T\) increases, this gap in "mistaken" pulls widens, leading to improved relative regret reduction. The asymptotic improvement ratio converges to \(1 - \sigma_{\max}^2\). In our Bernoulli experiment, arm variances \(p(1-p)\) for \(p \sim \mathrm{Uniform}(0.8, 0.95)\) lie in \([0.05, 0.16]\), giving \(\sigma_{\max}^2 \approx 0.16\), so the regret reduction is approximately \(1 - \sigma_{\max}^2 = 0.84\), i.e., \(84\%\) regret reduction, which aligns with both theory and empirical results in Figure~\ref{fig:regret_scaling}.
\subsection{Parameter Selection}
\label{Parameter Selection}
Real-world environments are complex and may not match ideal non-stationarity types. Practitioners can use these parameter ranges as a starting point and employ automated hyperparameter optimization techniques, such as Hyperband \cite{li2017hyperband} or bandit-based optimization \cite{huang2020asymptotically}, to fine-tune \(\alpha_0\) and \(\beta_0\) based on observed performance \cite{jamieson2016non}.

\begin{figure}[htbp]
    \centering
    \includegraphics[width=0.9\textwidth]{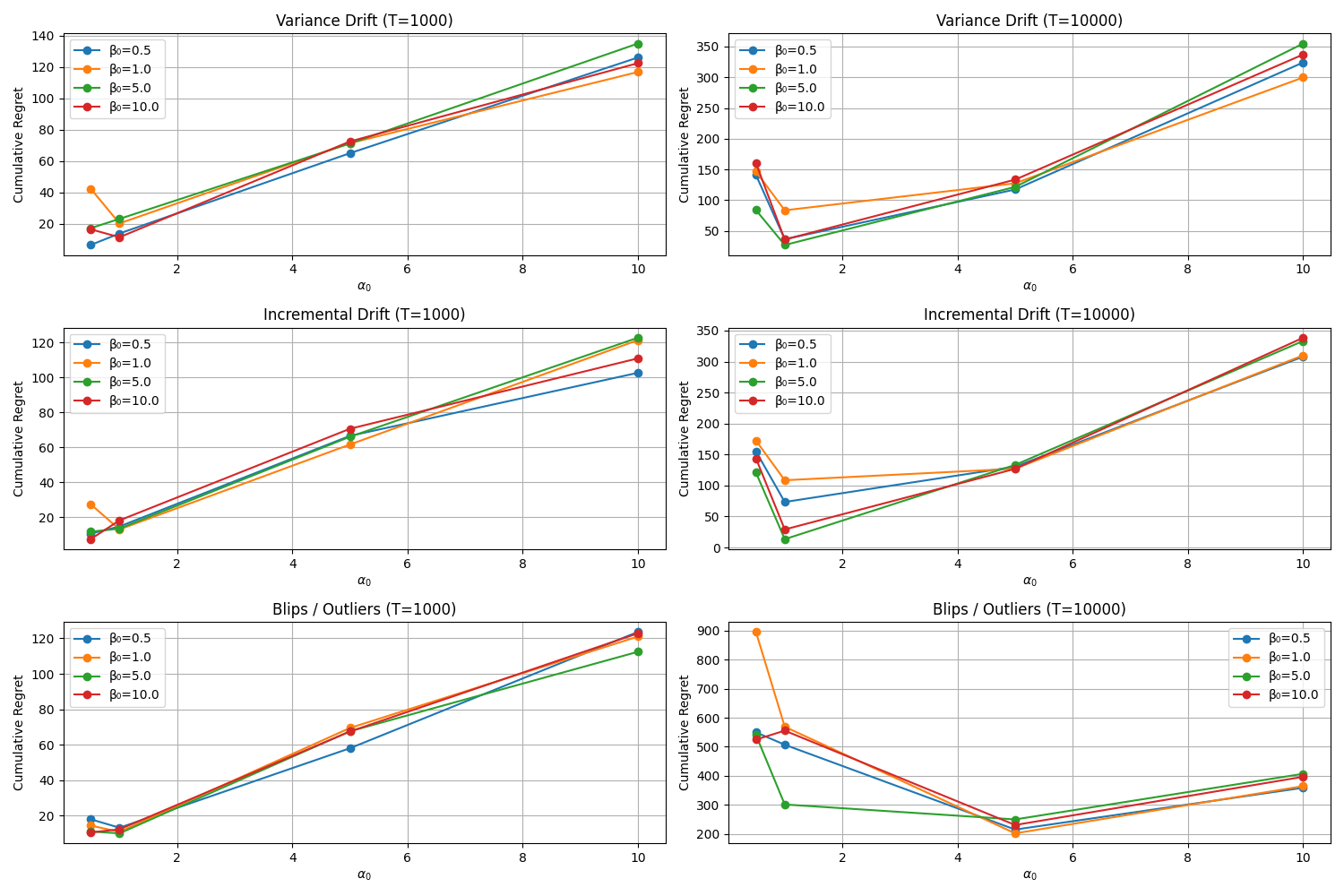}
    \caption{Cumulative Regret versus \(\alpha_0\) for different \(\beta_0\) values}
    \label{fig:regret_plot}
\end{figure}

We choose three non-stationarity types from Table \ref{tab:ns_mab_reclass_en} to conduct experiments and offer practical guidelines for deploying our algorithm in real-world non-stationary MAB scenarios. The parameter search results for the RAVEN-UCB algorithm, visualized in Figure \ref{fig:regret_plot} through cumulative regret curves across \(\alpha_0\) values with varying \(\beta_0\), demonstrate consistent hyperparameter selection patterns across distinct non-stationary regimes and time horizons (\(T=1000\), \(T=10000\)).

In the \textbf{Variance Drift} scenario, optimal parameters transition from \((\alpha_0=0.5, \beta_0=0.5)\) with regret 6.55 at \(T=1000\) to \((\alpha_0=1.0, \beta_0=5.0)\) with regret 27.14 at \(T=10000\). The flat minimal curve for \(\beta_0=0.5\) at \(T=1000\) indicates conservative exploration in short-term high-variance settings, while the sharp regret decline for \(\beta_0=5.0\) at \(T=10000\) validates enhanced stability through sustained exploration. For \textbf{Incremental Drift}, the \(T=1000\) optimum \((\alpha_0=0.5, \beta_0=10.0)\) yields regret 7.55 with downward-trending \(\beta_0=10.0\) curves, contrasting with the balanced \((\alpha_0=1.0, \beta_0=5.0)\) configuration (regret 13.48) dominating at \(T=10000\). In \textbf{Blips/Outliers} environments, short-term optimality at \(T=1000\) is achieved with \((\alpha_0=1.0, \beta_0=5.0)\) (regret 9.95), whereas long-term performance peaks at \(T=10000\) with \((\alpha_0=5.0, \beta_0=1.0)\) (regret 202.37), where the \(\beta_0=1.0\) curve dips at elevated \(\alpha_0\) values.

Observing Figure \ref{fig:regret_plot}, a clear pattern for choosing hyperparameters appear: (1) \(\alpha_0\) scales with time horizon length, increasing from 0.5 (short-term) to 5.0 (long-term) to balance exploration duration; (2) \(\beta_0\) inversely correlates with environmental volatility, with lower values (0.5–1.0) stabilizing high-variance regimes and higher values (5.0–10.0) smoothing gradual drifts. This structured parameter adaptation is very different from passive methods \cite{garivier2011upper} like Discounted UCB and Sliding Window UCB. These require manual tuning of discount factors or window sizes across scenarios. RAVEN-UCB's mechanism can reduce hyperparameter sensitivity. This is shown by the consistent regret trends across \(\beta_0\) values for each non-stationarity type.

In high-variance scenarios like traffic delay management, where delays vary due to peak hours or incidents, the preference for \(\beta_0 = 5.0\) with \(\alpha_0 = 1.0\) suggests that focusing on variance control over extended periods optimizes route selection, minimizing regret. For gradual changes, such as user preference shifts in recommendation systems, the optimal \(\alpha_0 = 1.0\), \(\beta_0 = 5.0\) supports a good strategy at \(T=10000\) , adjustable to \(\alpha_0 = 0.5\), \(\beta_0 = 10.0\) for short-term (e.g., daily) recommendations to quickly adapt to new preferences. In long-tail reward scenarios, such as inventory management during promotional spikes, the shift to \(\alpha_0 = 5.0\), \(\beta_0 = 1.0\) emphasizes aggressive exploration to capture rare high rewards without over-penalizing variance, aligning with the plot's trend for low \(\beta_0\).

\subsection{Simulation Study——Logistics Optimization Scenario}

In modern logistics and supply chain management, companies must make decisions under uncertainty. This makes the environment unpredictable. Traditional decision-making algorithms often struggle in such settings. So, companies need adaptive methods to quickly respond to changes. This scenario is highly relevant to several domains within logistics and supply chain systems. Logistics robotics and port operations require adaptive algorithms to manage dynamic operational constraints and demand surges \cite{ bierwirth2015handbook}. We can model these scenarios as a non-stationary multi-armed bandit problem. Here, each warehouse is an arm, and the reward is the delivery efficiency score. 

We model a logistics optimization scenario with \(K\) warehouses (arms), where the efficiency score of assigning an order to a warehouse is drawn from a normal distribution \(\mathcal{N}(\mu, \sigma^2)\). The range of means \([\mu_{\min}, \mu_{\max}]\) represents realistic variations in warehouse performance, where higher values indicate faster and more cost-effective operations. The variance range \([\sigma^2_{\min}, \sigma^2_{\max}]\) captures the uncertainty inherent in logistics, such as unpredictable delays or variable processing times. Every \(R\) time steps (representing customer orders), the means and variances of approximately one-third of the warehouses (\(\frac{K}{3}\) arms) are randomly reset to new values within their respective ranges using a uniform distribution. 
\begin{table}[H]
\centering
\caption{Simulation parameters and values used in the experiment}
\begin{tabular}{lll}
\toprule
\textbf{Symbol} & \textbf{Description} & \textbf{Value used in experiment} \\
\midrule
\(K\) & Warehouses & 100 \\
\(\mu_{\min}, \mu_{\max}\) & Range of mean efficiency scores & 0.3, 0.8 \\
\(\sigma^2_{\min}, \sigma^2_{\max}\) & Range of variances & 0.01, 0.09 \\
\(R\) & Shift interval  & 5,000 \\
\(T\) & Time steps & 50,000 \\
\(N\) & Number of independent trials & 50 \\

\bottomrule
\end{tabular}
\end{table}
This setup mimics real-world operational changes, such as traffic delays, inventory restocking, or weather-induced disruptions, which are common in logistics systems \cite{lee2018smart}. What's more, peak traffic hours, inventory updates, or seasonal weather patterns will also result in these situations, which are critical challenges in logistics optimization \cite{crainic2009models}. The simulation runs for \(T\) time steps, with \(N\) independent trials to ensure statistical reliability, and uses a fixed random seed for reproducibility.

By comparing RAVEN-UCB with eight other established MAB algorithms (UCB \cite{auer2002finite}, UCB-V \cite{audibert_explorationexploitation_2009}, \(\epsilon\)-greedy \cite{sutton2018reinforcement}, Thompson Sampling \cite{thompson1933likelihood}, f-dsw TS (min) \cite{cavenaghi_non_2021}, WLS + Optimistic TS \cite{chapelle2011empirical}, CCB \cite{liu_modification_2016}, and UCB-Imp \cite{abbasi2011improved}) in this logistics-inspired setting, we aim to demonstrate its effectiveness in dynamic decision-making environments. This research builds upon prior work in applying MAB algorithms to logistics optimization \cite{ivanov2019impact} and extends it by focusing on non-stationary environments \cite{slivkins2019introduction}.
\begin{table}[htbp]
    \centering
    \caption{Simulation Results}
    \label{tab:sim_results}
    \begin{tabular}{@{}lccc@{}}
        \toprule
        Algorithm          & Cumulative Reward & Cumulative Regret & Suboptimal Pulls (per trial) \\
        \midrule
        RAVEN-UCB          & 38276.8           & 1717.8            & 40824.2                  \\
        UCB                & 34551.2           & 5446.1            & 46595.1                  \\
        UCB-V              & 33985.8           & 6011.4            & 46475.4                  \\
        \(\epsilon\)-greedy & 36595.1           & 3397.8            & 43147.7                  \\
        Thompson Sampling  & 37029.8           & 2956.9            & 46254.5                  \\
        f-dsw TS (min)     & 34701.3           & 5305.7            & 48003.3                  \\
        WLS + Optimistic TS & 36224.0         & 3771.3            & 46357.8                  \\
        CCB                & 37306.9           & 2690.5            & 41095.1                  \\
        UCB-Imp            & 28192.8           & 11798.9           & 49446.5                  \\
        \bottomrule
    \end{tabular}
\end{table}
\begin{figure}[H]
\centering
\begin{subfigure}[t]{0.49\textwidth}
    \centering
    \includegraphics[width=\linewidth]{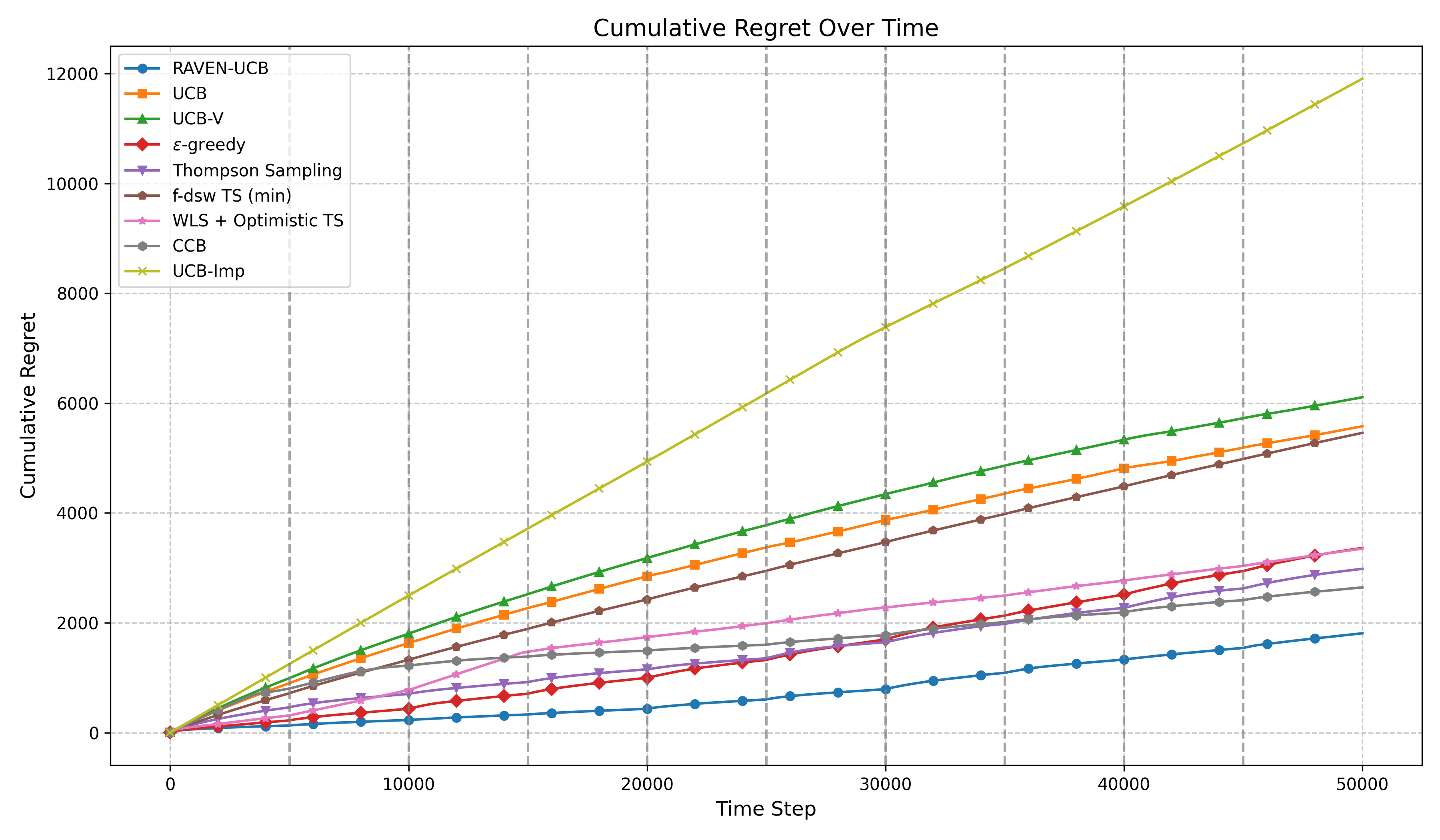}
    \caption{Cumulative regret over time}
    \label{fig:Cumulative regret over time}
\end{subfigure}
\hfill
\begin{subfigure}[t]{0.49\textwidth}
    \centering
    \includegraphics[width=\linewidth]{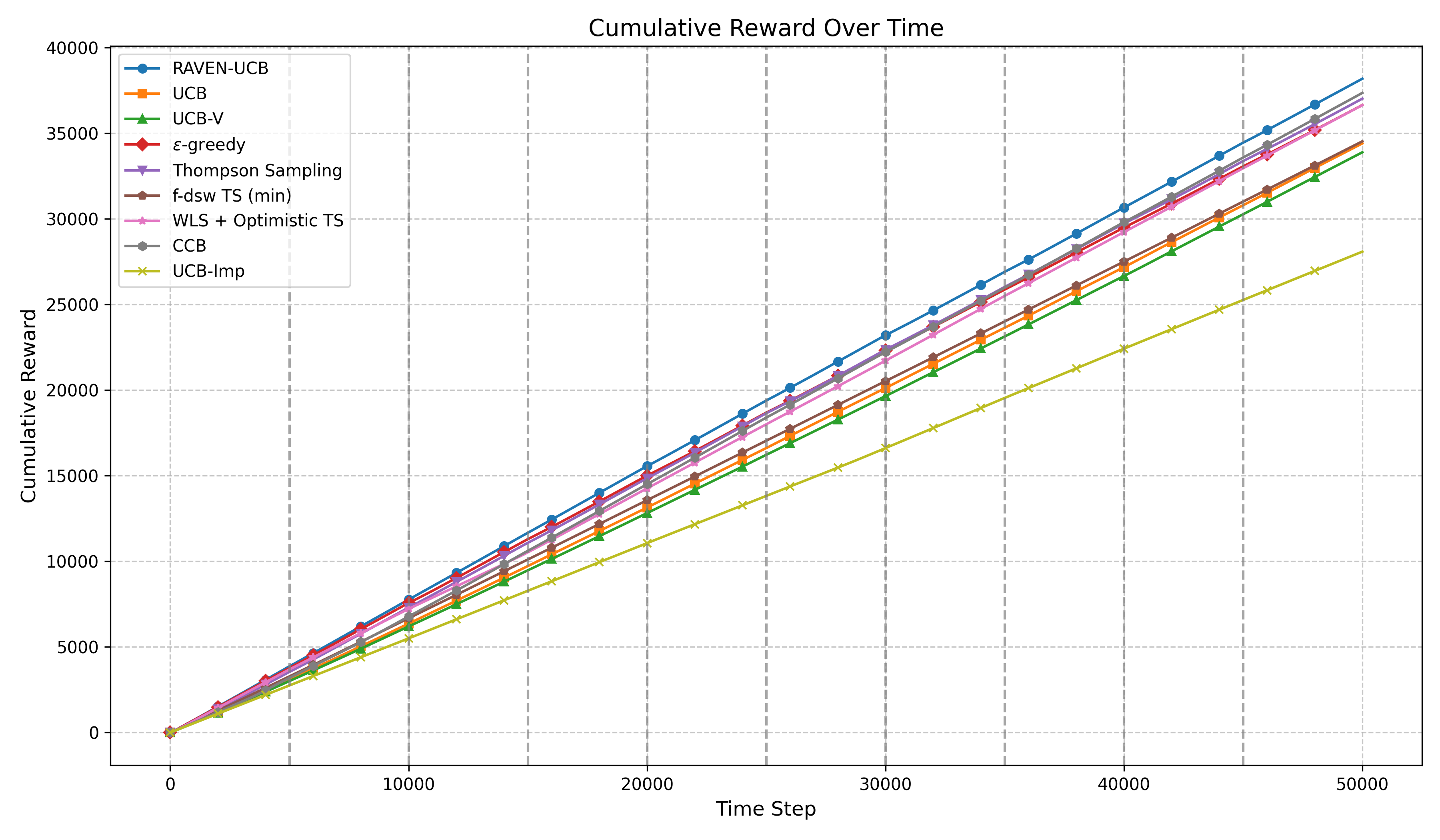}
    \caption{Cumulative reward over time}
    \label{fig:Cumulative reward over time}
\end{subfigure}
\caption{(a) Cumulative regret, (b) umulative reward}
\label{fig:regret_success}
\end{figure}
The results, averaged over 50 independent trials, are in Table \ref{tab:sim_results}. RAVEN-UCB achieves the highest cumulative reward of 38,276.8 and the lowest cumulative regret of 1,717.8 among all compared algorithms. The next best algorithm, CCB, records a cumulative reward of 37,306.9 and a regret of 2,690.5. In contrast, the standard UCB algorithm exhibits a significantly higher regret of 5,446.1. See Figure \ref{fig:Cumulative regret over time} and Figure \ref{fig:Cumulative reward over time}.
\begin{figure}[H]
    \centering
    \includegraphics[width=1.0\linewidth]{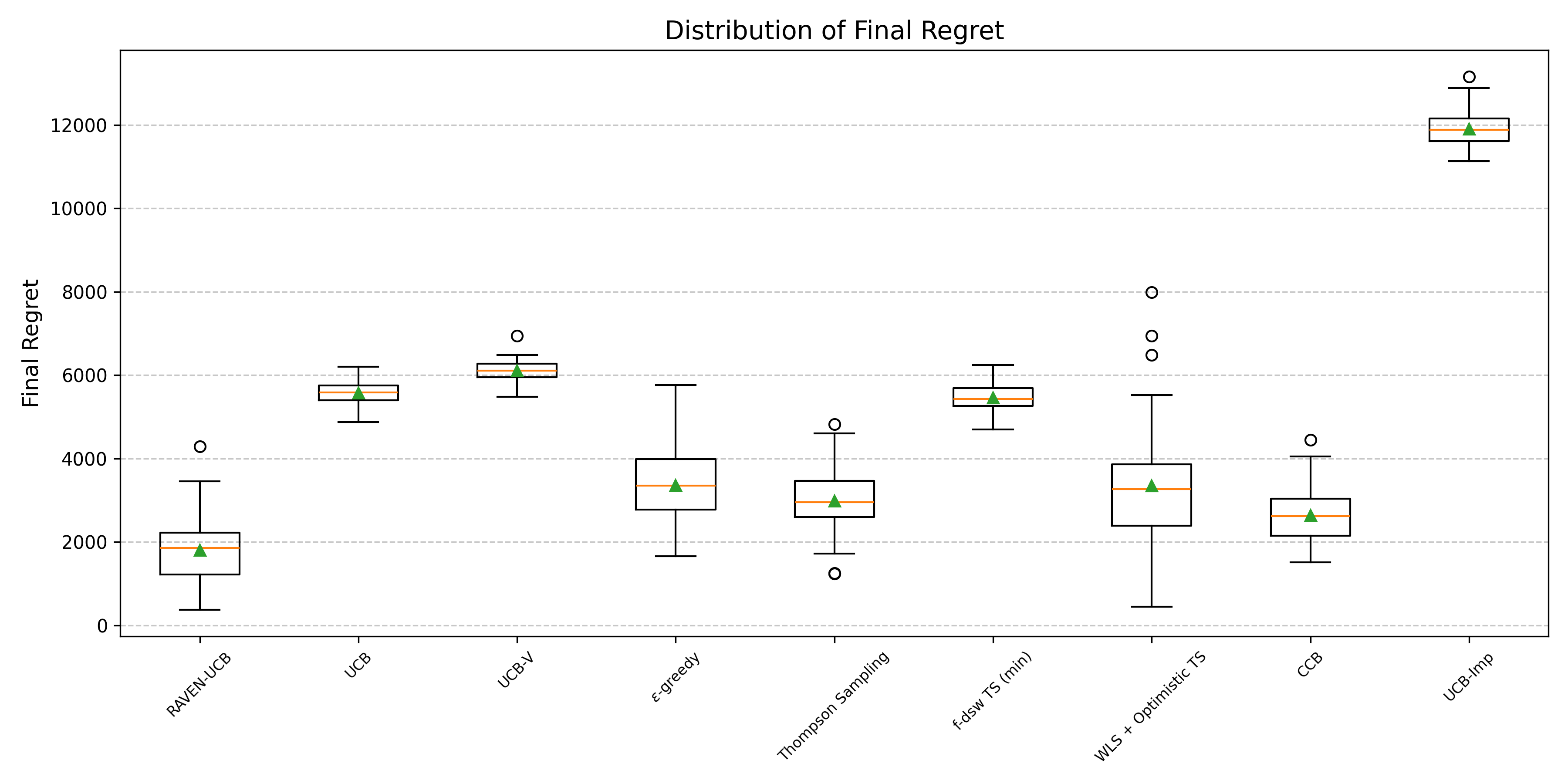}
    \caption{Boxplots of the results}
    \label{fig:boxplots of the results}
\end{figure}
RAVEN-UCB performs well because it uses a variance-adaptive exploration strategy. This helps it quickly find and adapt to changes in warehouse efficiency.  This is particularly valuable in logistics, where rapid adaptation to operational changes is critical. But using a normal distribution for rewards fits continuous efficiency metrics. It may not be the best choice where binary outcomes (e.g., on-time delivery rates) are more relevant.

\section{Conclusion}
\label{Conclusion}
In this paper, we introduced and analyzed the RAVEN-UCB algorithm. It addresses non-stationarity in Multi-Armed Bandit (MAB) problems through a variance-adaptive approach. This dynamically balances exploration and exploitation. We review the MAB framework and its uses in online advertising, recommendation systems, and logistics optimization. We classifying non-stationarity into distributional parameter changes (DPC), periodic changes (PC), and temporary fluctuations (TF). The RAVEN-UCB algorithm uses a new approach with a variance-based exploration signal. It uses the sample variance to guide exploration, and employs a logarithmically decaying coefficient $\alpha_t = \frac{\alpha_0}{\log(t + \epsilon)}$ to adjust exploration intensity. Computational efficiency improves with recursive formulas, reducing the time complexity to $\mathcal{O}(1)$. Theoretically, the algorithm achieves gap-dependent regret $\mathcal{O}\left(\frac{K \sigma_{\max}^2 \log T}{\Delta}\right)$ and gap-independent regret $\mathcal{O}(\sqrt{K T \log T})$. We conducted many experiments to evaluate the RAVEN-UCB algorithm: 

Firstly, in regret performance experiments with sub-Gaussian rewards, RAVEN-UCB achieved $1-\sigma_{\max} ^2$ regret reduction with Bernoulli rewards. This matches  its gap-dependent regret bound. It outperforms UCB1’s $\mathcal{O}\left(\frac{K \log T}{\Delta}\right)$ much better.

Secondly, in the parameter selection experiments across three representative non-stationary scenarios, grid search over \(\alpha_0\) and \(\beta_0\) helped with practical tuning strategies. These findings provide actionable guidance for using RAVEN-UCB in dynamic environments and support stable performance across varying types of non-stationarity.

Thirdly, in a simulated logistics optimization scenario with $K=100$ warehouses as arms and rewards drawn from a normal distribution $\mathcal{N}(\mu, \sigma^2)$, RAVEN-UCB recorded a cumulative regret of 1717.8 over $T=50,000$ steps and $N=50$ trials, far surpassing UCB1’s 5446.1, demonstrating its ability to adapt to changing efficiency distributions via a variance-adaptive strategy.

Based on our experiments, we provide the following practical guidance for parameter selection in RAVEN-UCB: 

(1) In high-variance environments (e.g., traffic delay management), setting \(\alpha_0 = 1.0\), \(\beta_0 = 5.0\) helps stabilize exploration over long horizons. 

(2) For gradual changes (e.g., evolving user preferences), \(\alpha_0 = 1.0\), \(\beta_0 = 5.0\) is effective for long-term adaptation, while \(\alpha_0 = 0.5\), \(\beta_0 = 10.0\) works better for short-term responsiveness. 

(3) In rare-event or outlier-driven contexts (e.g., promotional spikes in inventory management), aggressive exploration with \(\alpha_0 = 5.0\), \(\beta_0 = 1.0\) captures high-reward opportunities efficiently. 

Future work could explore applying RAVEN-UCB to real-world datasets in domains such as e-commerce, traffic management, or financial decision-making, where non-stationarity is inherent and variance dynamics are complex. This would help validate its robustness and scalability beyond controlled simulations. Additionally, another direction is to investigate automated adaptation mechanisms for hyperparameter tuning, potentially leveraging meta-learning or reinforcement meta-bandits.
Also, integrating RAVEN-UCB with contextual bandit frameworks to incorporate side information may boost its performance in diverse environments.

\section*{Declarations}

\begin{itemize}
\item Data availability: This study does not use external datasets. All data were generated through simulations as described in Section \ref{Experiments}. The data can be made available upon reasonable request from the corresponding author.
\item Code availability: The code used to implement the RAVEN-UCB algorithm and conduct the simulations in Sections \ref{Experiments} is available at \url{https://github.com/66661654/Raven-UCB}.
\item Author contribution: Conceptualization, J. Fang and Yuxun Chen; Methodology, J. Fang and Yuxun Chen; Coding, J. Fang; Validation, Yuxun Chen; Writing—original draft, J. Fang and Yuxun Chen; Writing—review and editing, Yuxun Chen, Yuxin Chen, and C. Zhang. All authors have read and agreed to the published version of the manuscript.
\end{itemize}

\appendix
\section{Appendix}
\label{app:Appendix}

\subsection{Derivation of Recursive Formulas for Sample Mean and Variance}
\label{app:recursive_derivation}
Upon the arrival of a new observation \( x_{n+1} \), the sample mean for \( n+1 \) observations, \( \overline{x}_{n+1} \), is similarly defined:
\begin{equation}
\overline{x}_{n+1} = \frac{1}{n+1} \sum_{i=1}^{n+1} x_i.
\label{eq:mean_nplus1}
\end{equation}
\begin{equation}
\sum_{i=1}^{n+1} x_i = \sum_{i=1}^{n} x_i + x_{n+1} = n \overline{x}_n + x_{n+1}.
\label{eq:sum_split}
\end{equation}
So, we get (\ref{eq:recursive_mean}):
\begin{equation}
\overline{x}_{n+1} = \overline{x}_n + \frac{x_{n+1} - \overline{x}_n}{n+1}.
\label{eq:recursive_mean_app}
\end{equation}
Turning to the sample variance, we adopt the unbiased estimator, which accounts for the degrees of freedom lost in estimating the sample mean. For \( n \) observations, the sample variance \( S_n^2 \) is defined as:
\begin{equation}
S_n^2 = \frac{1}{n-1} \sum_{i=1}^{n} (x_i - \overline{x}_n)^2, \quad \text{for} \quad n \geq 2.
\label{eq:sample_variance_app}
\end{equation}
Let us define \( M_n = \sum_{i=1}^{n} (x_i - \overline{x}_n)^2 \), so that
\begin{equation}
S_n^2 = \frac{M_n}{n-1}.
\label{eq:Mn_variance}
\end{equation}
\begin{equation}
S_{n+1}^2 = \frac{1}{n} \sum_{i=1}^{n+1} (x_i - \overline{x}_{n+1})^2 = \frac{M_{n+1}}{n},
\label{eq:sample_variance_nplus1}
\end{equation}
where
\begin{equation}
M_{n+1} = \sum_{i=1}^{n+1} (x_i - \overline{x}_{n+1})^2.
\label{eq:Mnplus1_def}
\end{equation}
\begin{equation}
M_{n+1} = \sum_{i=1}^{n} (x_i - \overline{x}_{n+1})^2 + (x_{n+1} - \overline{x}_{n+1})^2.
\label{eq:Mnplus1_split}
\end{equation}
We need to express \( x_i - \overline{x}_{n+1} = (x_i - \overline{x}_n) - (\overline{x}_{n+1} - \overline{x}_n) \), noting that \( \sum_{i=1}^{n} (x_i - \overline{x}_n) = 0 \). Summing over \( i = 1 \) to \( n \), of course, using (\ref{eq:recursive_mean}):
\begin{equation}
\sum_{i=1}^{n} (x_i - \overline{x}_{n+1})^2 = M_n + n (\overline{x}_{n+1} - \overline{x}_n)^2.
\label{eq:Mnplus1_part1}
\end{equation}
Next, compute the contribution of the new observation,
\begin{equation}
x_{n+1} - \overline{x}_{n+1} = x_{n+1} - \left( \overline{x}_n + \frac{x_{n+1} - \overline{x}_n}{n+1} \right) = \frac{n (x_{n+1} - \overline{x}_n)}{n+1},
\label{eq:xnplus1_deviation}
\end{equation}
so:
\begin{equation}
(x_{n+1} - \overline{x}_{n+1})^2 = \left( \frac{n}{n+1} (x_{n+1} - \overline{x}_n) \right)^2 = \frac{n^2}{(n+1)^2} (x_{n+1} - \overline{x}_n)^2.
\label{eq:xnplus1_deviation_sq}
\end{equation}
Additionally,
\begin{equation}
(\overline{x}_{n+1} - \overline{x}_n)^2 = \left( \frac{x_{n+1} - \overline{x}_n}{n+1} \right)^2 = \frac{1}{(n+1)^2} (x_{n+1} - \overline{x}_n)^2.
\label{eq:mean_diff_sq}
\end{equation}
Therefore:
\begin{equation}
M_{n+1} = M_n + n \cdot \frac{1}{(n+1)^2} (x_{n+1} - \overline{x}_n)^2 + \frac{n^2}{(n+1)^2} (x_{n+1} - \overline{x}_n)^2 = M_n + \frac{n}{n+1} (x_{n+1} - \overline{x}_n)^2.
\label{eq:Mnplus1_final}
\end{equation}
Hence, the updated sample variance is:
\begin{equation}
S_{n+1}^2 = \frac{M_{n+1}}{n} = \frac{M_n}{n} + \frac{1}{n+1} (x_{n+1} - \overline{x}_n)^2.
\label{eq:var_update_step1}
\end{equation}
Since \( M_n = (n-1) S_n^2 \), we have \( \frac{M_n}{n} = \frac{(n-1) S_n^2}{n} = \left(1 - \frac{1}{n}\right) S_n^2 \), and recognizing that \( (x_{n+1} - \overline{x}_n)^2 = (n+1)^2 (\overline{x}_{n+1} - \overline{x}_n)^2 \), the variance update can be expressed as (\ref{eq:recursive_variance}):
\begin{equation}
S_{n+1}^2 = \left(1 - \frac{1}{n}\right) S_n^2 + (n+1) (\overline{x}_{n+1} - \overline{x}_n)^2.
\label{eq:var_update_final}
\end{equation}
Each operation executes in constant time, independent of \( n \), yielding a time complexity of \( O(1) \) per update. In contrast, the naive approach recomputes the mean and variance from all \( n+1 \) observations, requiring \( O(n) \) time per update, with a total complexity of \( O(n^2) \) over \( n \) updates.

\subsection{Proof of the Regret Upper Bound}
\label{app:regret_bound}
Consider a multi-armed bandit with $K$ arms. Each arm $k$ has rewards $X_k$ with mean $\mu_k$ and sub-Gaussian parameter $\sigma_k^2$. The optimal arm is $k^* = \arg\max_k \mu_k$ with mean $\mu^* = \mu_{k^*}$. The gap for a suboptimal arm $k \neq k^*$ is $\Delta_k = \mu^* - \mu_k > 0$, and $\Delta = \min_{k \neq k^*} \Delta_k$. The cumulative regret over $T$ rounds is :\cite{auer2002finite}:
\begin{equation}
R(T) = \mathbb{E}\left[ \sum_{t=1}^T (\mu^* - \mu_{k_t}) \right] = \sum_{k \neq k^*} \Delta_k \cdot \mathbb{E}[N_k(T)],
\label{eq:regret}
\end{equation}
where $N_k(T)$ is the number of pulls of arm $k$.

The upper confidence bound \( U_k(t) \) typically includes the empirical mean \( \hat{\mu}_k(t) \) and an exploration term. For simplicity, we define \( U_k(t) = \hat{\mu}_k(t) + c_k(t) \), where the confidence radius \( c_k(t) \) is given in Algorithm~\ref{alg:v_ucb}.

\begin{equation}
c_k(t) = \alpha_t \cdot \sqrt{\frac{\ln(t )}{N_k(t) + 1}} + \beta_0 \cdot \sqrt{\frac{\hat{\sigma}_k^2}{N_k(t) + 1}} + \epsilon,
\end{equation}
where \( \alpha_t = \alpha_0 / \log(t + \epsilon) \). When \( T \) is large, the term \( \epsilon \) becomes negligible. Moreover, the empirical variance \( \hat{\sigma}_k^2 \) converges to the true variance \( \sigma_k^2 \).
Suppose \( \alpha_0 \sqrt{\ln t} + \beta_0 \sigma_k = 2 \sigma_k \sqrt{\ln t} \), the confidence radius can be simplified to:
\begin{equation}
c_k(t) = \sqrt{\frac{4 \sigma_k^2 \log t}{N_k(t)}},
\end{equation}
For sub-Gaussian rewards we use the concentration inequality
\cite{boucheron_concentration_2004}
:
\begin{equation}
\mathbb{P}(|\hat{\mu}_k(t) - \mu_k| \geq \varepsilon) \leq 2 \exp\left( -\frac{N_k(t) \varepsilon^2}{2 \sigma_k^2} \right).
\label{eq:concentration}
\end{equation}
Define $E_k(t) = \{ k_t = k \}$. Then $E_k(t) \subseteq A_t \cup B_t$, where $A_t = \{ U_{k^*}(t) < \mu^* \}$, $B_t = \{ U_k(t) \geq \mu_k + \frac{\Delta_k}{2} \}$, so:
\begin{equation}
\mathbb{P}(E_k(t)) \leq \mathbb{P}(A_t) + \mathbb{P}(B_t).
\label{eq:union}
\end{equation}
For $A_t$:
\begin{equation}
\mathbb{P}(A_t) = \mathbb{P}(\hat{\mu}_{k^*}(t) - \mu^* < -c_{k^*}(t)) \leq \exp\left( -\frac{N_{k^*}(t) c_{k^*}(t)^2}{2 \sigma_{k^*}^2} \right) = \frac{1}{t^2},
\label{eq:prob_A}
\end{equation}
since $c_{k^*}(t)^2 = \frac{4 \sigma_{k^*}^2 \log t}{N_{k^*}(t)}$. For $B_t$, set $\varepsilon = \frac{\Delta_k}{2} - c_k(t)$. If $N_k(t) \geq \frac{16 \sigma_k^2 \log t}{\Delta_k^2}$, then $c_k(t) \leq \frac{\Delta_k}{2}$, and:
\begin{equation}
\mathbb{P}(B_t) = \mathbb{P}\left( \hat{\mu}_k(t) - \mu_k \geq \frac{\Delta_k}{2} - c_k(t) \right) \leq \exp\left( -\frac{N_k(t) \left( \frac{\Delta_k}{2} \right)^2}{2 \sigma_k^2} \right) \leq \frac{1}{t^2}.
\label{eq:prob_B}
\end{equation}
Use (\ref{eq:union}), $\mathbb{P}(E_k(t)) \leq \frac{2}{t^2}$. Set $n_k = \frac{16 \sigma_k^2 \log T}{\Delta_k^2}$. Noting the result of Basel Problem \cite{risch1970solution,moreno2016short}, the expected sub-pulls are : \(\mathbb{E}[N_k(T)] \leq n_k + \sum_{t = n_k + 1}^{T} \frac{2}{t^2}<n_k + \frac{\pi^2}{6} = \mathcal{O}\left( \frac{\sigma_k^2 \log T}{\Delta_k^2} \right)\).

The gap-dependent regret is:
\begin{equation}
\mathbb{E}[R(T)] < \sum_{k \neq k^*} \frac{16 \sigma_k^2 \log T}{\Delta_k} \leq \mathcal{O}\left( \frac{K \sigma_{\max}^2 \log T}{\Delta} \right).
\label{eq:gap_dep}
\end{equation}
For the gap-independent bound, assume $\mathbb{E}[N_k(T)] \approx l$, with:
\begin{equation}
c_k(t) = \sqrt{\frac{4 \sigma_k^2 \log T}{l}}, \quad \Delta_k \approx c_k(t).
\label{eq:delta_l}
\end{equation}
By Cauchy-Schwarz on \eqref{eq:regret}:
\begin{equation}
\mathbb{E}[R(T)] \leq \sqrt{(K-1) \Delta_k^2} \cdot \sqrt{(K-1) l^2}.
\label{eq:cauchy}
\end{equation}
Substitute \eqref{eq:delta_l}:
\begin{equation}
\mathbb{E}[R(T)] \approx \sqrt{4 \sigma_k^2 (K-1) \log T \cdot l}.
\label{eq:regret_l}
\end{equation}
With $\sum_{k=1}^K N_k(T) = T$, set $m = (K-1) l \approx \sqrt{K T \log T}$, so:
\begin{equation}
l \approx \sqrt{\frac{T \log T}{K}}, \quad \mathbb{E}[R(T)] \approx \sqrt{4 \sigma_{\max}^2 K T \log T} = \mathcal{O}(\sqrt{K T \log T}).
\label{eq:gap_indep}
\end{equation}

\newpage
\printbibliography
\end{document}